\documentclass{article}

\PassOptionsToPackage{numbers, compress}{natbib}

 \usepackage[preprint]{nips_2018}

\usepackage[utf8]{inputenc} 
\usepackage[T1]{fontenc}    
\usepackage{hyperref}       
\usepackage{url}            
\usepackage{amsfonts}       
\usepackage{times}
\usepackage{graphicx} 
\usepackage{amsmath} 
\usepackage{amssymb}
\usepackage{booktabs,multirow}
\usepackage[bf,footnotesize]{caption}
\usepackage{tabularx}
\usepackage[]{verbatim}
\usepackage{algorithm}
\usepackage{algorithmic}
\usepackage{mathtools}
\usepackage{enumitem}
\usepackage{color}
\usepackage{balance}
\usepackage{url}
\usepackage{xcolor}
\usepackage{epsfig}
\usepackage{wrapfig}
\usepackage[export]{adjustbox}
\graphicspath{{./}{./figures/}}

\def\Mat#1{{\boldsymbol{#1}}}
\newcommand{\mycomment}[1]{}

\title{Learning Factorized Representations\\ for Open-set Domain Adaptation}

\author{
  Mahsa Baktashmotlagh \\
  QUT\\
  \And
    Masoud Faraki \\
  Monash University\\
    \And
    Tom Drummond \\
  Monash University\\
    \And
    Mathieu Salzmann \\
  EPFL\\
}

\begin{document}

\maketitle

\begin{abstract}
Domain adaptation for visual recognition has undergone great progress in the past few years. Nevertheless, most existing methods work in the so-called closed-set scenario, assuming that the classes depicted by the target images are exactly the same as those of the source domain. In this paper, we tackle the more challenging, yet more realistic case of open-set domain adaptation, where new, unknown classes can be present in the target data. While, in the unsupervised scenario, one cannot expect to be able to identify each specific new class, we aim to automatically detect which samples belong to these new classes and discard them from the recognition process. To this end, we rely on the intuition that the source and target samples depicting the known classes can be generated by a shared subspace, whereas the target samples from unknown classes come from a different, private subspace. We therefore introduce a framework that factorizes the data into shared and private parts, while encouraging the shared representation to be discriminative. Our experiments on standard benchmarks evidence that our approach significantly outperforms the state-of-the-art in open-set domain adaptation.
\end{abstract}


%
\section{Introduction}
In many practical machine learning scenarios, the test samples are drawn from a different distribution from the training ones, due to varying acquisition conditions, such as different data sources, illumination conditions and cameras, in the context of visual recognition. Over the years, great progress has been achieved to tackle this problem, known has the domain shift. In particular, many methods aim to align the source (i.e., training) and target (i.e., test) distributions by learning domain-invariant embeddings~\citep{pan2011domain,gong2012geodesic,fernando2013unsupervised,sun2016return}, the most recent approaches relying on deep networks~\citep{backprop,long2015learning,bousmalis2016domain,tzeng2017adversarial,long2016deep,yan2017mind}.

While effective, these methods work under the assumption that the source and target data contain exactly the same classes. In practice, however, this assumption may easily be violated, as the target data will often contain additional classes that were not present within the source data. For example, when training a model to recognize office objects from images, as with the popular Office dataset~\citep{saenko2010adapting}, one should still expect to see new objects, unobserved during training, when deploying the model in the real world. While one should not expect the model to recognize the specific class of such objects, at least in unsupervised domain adaptation where no target labels are provided, it would nonetheless be beneficial to identify these objects as unknown instead of misclassifying them. This was the task addressed by~\cite{busto2017open} in their so-called \emph{open-set} domain adaptation approach. This method aims to learn a mapping from the source samples to a subset of the target ones corresponding to those identified as coming from known classes. While reasonably effective, this procedure involves alternatively solving for the mapping and the assignment of source and target samples to known/unknown classes, which, as discussed in our experiments, can be relatively costly. 

In this paper, we introduce a novel approach to open-set domain adaptation based on learning a factorized representation of the source and target data. In essence, we seek to model the samples from the known classes with a low-dimensional subspace, shared by the source and target domains, and the target samples from unknown classes with another subspace, specific to the target domain. We then make use of group sparsity to encourage each target sample to be reconstructed by only one of these subspaces, which in turns lets us identify if this sample corresponds to a known or unknown class. We further show that we can obtain a more discriminative shared representation by jointly learning a linear classifier within our framework. Ultimately, our approach therefore allows us to jointly separate the target samples between known and unknown classes and represent the source and target samples within a consistent, shared latent space.

We demonstrate the effectiveness of our approach on several open set domain adaptation benchmarks for visual object recognition. Our method consistently and significantly outperforms the state-of-the-art technique of~\cite{busto2017open}, thus showing the benefits of learning shared and private representations corresponding to the known and unknown classes, respectively. Furthermore, it is faster than the algorithm of~\cite{busto2017open} by an order of magnitude.

\section{Related Work}
\label{related}
\vspace{-0.2cm}
Domain adaptation for visual recognition has become increasingly popular over the past few years, in large part thanks to the benchmark Office dataset of~\cite{saenko2010adapting}. A natural approach to tackling the domain shift consists of learning a transformation of the data such that the distributions of the source and target samples are as similar as possible in the resulting space. In practice, this is achieved by minimizing a distance between these distributions, such as the Maximum Mean Discrepancy (MMD) in~\citep{mb1}, the Hellinger distance in~\citep{mb2}, or a distance based on second-order statistics in~\citep{sun2016return}. Instead of learning a transformation of the data, other methods have been proposed to re-weight the source samples, so as to rely more strongly on those that look similar to the target ones~\citep{quinonerocovariate,gong2}.

With the advent of deep learning for visual recognition, domain adaptation research also eventually turned to exploiting deep networks. While it was initially shown that deep features were more robust than handcrafted ones to the domain shift~\citep{donahue2013decaf}, translating the above-mentioned distribution-matching ideas, such as the use of (variations of) the MMD~\citep{tzeng2014deep,long2015learning,long2016unsupervised,rozantsev2018pami} or of second-order statistics~\citep{sun2016deep}, to end-to-end learning proved even more effective. In this context, other ideas were  introduced, such as learning intermediate representations to interpolate between the source and target domains~\citep{chopra,tzeng2015simultaneous}, the use of adversarial domain classifiers~\citep{backprop,tzeng2017adversarial}, and additional reconstruction loss terms~\citep{ghifary2016deep}.

Despite achieving great progress to tackle the domain shift, all the aforementioned methods are designed for the closed-set scenario, where the source and target data depict the exact same set of classes. Inspired by recent advances in open-set recognition~\citep{bendale2015towards,scheirer2014probability}, the work of~\cite{busto2017open} constitutes the first and only attempt to address the more realistic case where the target data contains samples from new, unknown classes. To achieve this,~\cite{busto2017open} proposed to jointly learn the assignments of the target samples to known/unknown classes and a mapping from the source data to the target samples depicting known classes. The resulting learning problem was solved by alternatively optimizing for the assignments and for the mapping, which can be costly.

Here, we introduce another solution to the open-set domain adaptation problem, where we model the source and target data with subspaces. Subspace-based representations have proven effective for domain adaptation~\citep{gong2012geodesic,gopalan,fernando2013unsupervised}. Here, however, we exploit them in a different manner, based on the intuition that source samples and target samples from the known classes can be generated by a shared subspace, whereas target samples from unknown classes come from a private subspace. While the notion of shared-private representations has been exploited in the past, e.g., for multiview learning~\citep{jia2010factorized} and for closed-set domain adaptation~\citep{bousmalis2016domain}, the resulting techniques all use them to encode each sample as a mixture of shared and private information. By contrast, here, we aim to model each target sample as being generated by either the shared subspace or the private one, which is crucial to identify the target samples depicting unknown classes.

Our experiments evidence that our open-set domain adaptation approach, based on shared-private representations, is more effective than the one of~\cite{busto2017open}, consistently outperforming it on several datasets and source/target pairs, and also faster by an order of magnitude. Note that, in contrast to~\cite{busto2017open} that exploits additional source data from unknown classes to improve accuracy, our method yields accurate results without such additional data, although it can also handle it. This, we believe, better reflects a practical scenario, where one can typically annotate all the source data.

\section{Our Approach}
\label{sec:proposed}
The key idea behind our formulation is to find low-dimensional representations of the data, factorized into a subspace shared by the source samples and the target ones coming from known classes and another subspace specific to the target samples from unknown classes. Note that, when referring to target samples from known classes, we do not mean that these samples are labeled, but rather that they belong to the same set of classes as the source data. As a matter of fact, throughout the paper, we focus on the unsupervised domain adaptation scenario, where no target annotations are provided. In the remainder of this section, we first introduce the optimization problem at the heart of our approach, and then discuss two extensions of this basic formulation.

\subsection{FRODA: Factorized Representations for Open-set Domain Adaptation}
Given $n_s$ source samples, grouped in a matrix $\Mat{X}_s \in \mathbb{R}^{D\times n_s}$ and $n_t$ target samples represented by $\Mat{X}_t \in \mathbb{R}^{D\times n_t}$, our goal is to estimate a low-dimensional representation of each sample, such that the source and target samples coming from the same classes are generated by a shared subspace, whereas the target data from new, unknown classes are generated by a different, specific subspace. To this end, let $\Mat{V} \in \mathbb{R}^{D \times d}$ be the matrix encoding the shared subspace, with $d \ll D$, and $\Mat{U} \in \mathbb{R}^{D \times d}$ the one representing the private subspace. A naive approach to finding the low-dimensional representations of the data would involve solving
\begin{align}
\min_{\Mat{U},\Mat{T},\Mat{V},\Mat{S}} & \hspace{2ex} \|\Mat{X}_t-\Mat{B}\Mat{T}\|_F^2 + \alpha \|\Mat{X}_s-\Mat{V}\Mat{S}\|_F^2\;,
\label{eqn:naive}
\end{align}
where $\alpha$ sets the relative influence of both terms, $\Mat{B} = [\Mat{V}, \Mat{U}] \in \mathbb{R}^{D \times 2d}$, and $\Mat{T}$ and $\Mat{S}$ encode the low-dimensional representations of the target and source data, respectively. This simple formulation, however, does not aim to separate the target samples belonging to known classes from the unknown ones, and thus will represent each target sample as a mixture of shared and private information.

Intuitively, we would rather like each target sample to be generated by either the shared subspace $\Mat{V}$, or the private one $\Mat{U}$. To address this, we propose to make use of a group sparsity regularizer on the coefficients of the target samples. Specifically, we split the coefficient vector $\Mat{T}_i$ for target sample $i$ into a part $\Mat{T}_i^v$ that corresponds to the shared subspace and a part $\Mat{T}_i^u$ that corresponds to the private one. We then encourage that either of these two parts goes to zero for each sample. To this end, we therefore write the optimization problem
\begin{align}
\min_{\Mat{U},\Mat{T},\Mat{V},\Mat{S}} & \hspace{2ex} \|\Mat{X}_t-\Mat{B}\Mat{T}\|_F^2 + \alpha \|\Mat{X}_s-\Mat{V}\Mat{S}\|_F^2+\lambda_1 \sum_{i=1}^{n_t}\left( \|\Mat{T}_i^v\| + \|\Mat{T}_i^u\|\right)\nonumber\\
 s.t.&  \hspace{2ex} \sum_{j=1}^{d}\|\Mat{U}_j\| \leq 1,\sum_{j=1}^{d}\|\Mat{V}_j\| \leq 1\;, &
\label{eqn:eq1}
\end{align}
where the constraints prevent the basis vectors of the subspaces from growing while the coefficients decrease~\cite{lee2007efficient}, and where $\lambda_1$ is a scalar controlling the strength of the group sparsity regularizer. In essence, this formulation allows each target sample to be reconstructed from either the shared subspace or the target one, which reduces the influence of the samples from unknown classes on learning the shared representation.

%
\vspace{-0.2cm}
\paragraph{Optimization.}
To solve~\eqref{eqn:eq1} efficiently, we alternatively update one variable at a time while keeping the other ones fixed. Below, we describe the different updates.

$\Mat{B}$-minimization: Given the coefficients $\Mat{S}$ and $\Mat{T}$, we update the tuple $(\Mat{U},\Mat{V})$ by solving 
\begin{align}
\min_{\Mat{U}} & \hspace{2ex} \|\Mat{A}-\Mat{U}\Mat{T}^u\|_F^2  \hspace{2ex}
 s.t.   \sum_{j=1}^{d}\|\Mat{U}_j\| \leq 1 \;,
\label{eqn:eq2}
\end{align}
with $\Mat{A}=\Mat{X}_t-\Mat{V}\Mat{T}^v$, and
\begin{align}
\min_{\Mat{V}} & \hspace{2ex} \|\Mat{A'}-\Mat{V}\Mat{T}^v\|_F^2 + \alpha \|\Mat{X}_s-\Mat{V}\Mat{S}\|_F^2 \hspace{2ex}
 s.t.   \sum_{j=1}^{d}\|\Mat{V}_j\| \leq 1 \;, 
\label{eqn:eqv}
\end{align}
with $\Mat{A'} = \Mat{X}_t-\Mat{U}\Mat{T}^u$.
\mycomment{
\begin{align}
\min_{\Mat{V}} & \hspace{2ex} \|\Mat{A'}-\Mat{V}\Mat{T'}\|_F^2  \hspace{2ex}
 s.t.   \sum_{j=1}^{d}\|\Mat{V}(j)\| <= 1 \; 
\label{eqn:eq}
\end{align}
with $\Mat{A'} = \begin{pmatrix}
\Mat{X}_t-\Mat{O}_t\Mat{T_o}\\ \sqrt{\alpha}\Mat{X_s} 
\end{pmatrix}$ and $\Mat{T'} = \begin{pmatrix}
\Mat{T_v} \\ \sqrt{\alpha}\Mat{S} 
\end{pmatrix}$.
}
These two sub-problems can be solved efficiently using the Lagrange dual formulation introduced in~\citep{lee2007efficient}.

$\Mat{T}$-minimization: Minimizing~\eqref{eqn:eq1} with respect to $\Mat{T}$ alone, with all the other parameters fixed, yields 
\begin{align}
\min_{\Mat{T}} & \hspace{2ex} \|\Mat{X}_t-\Mat{B}\Mat{T}\|_F^2 +\lambda_1 \sum_{i=1}^{n_t}\left( \|\Mat{T}_i^v\| + \|\Mat{T}_i^u\|\right)\;,
\label{eqn:eq4}
\end{align}
which can be solved efficiently via the proximal gradient method~\citep{mairal2014spams}. 

$\Mat{S}$-minimization: Solving~\eqref{eqn:eq1} with respect to $\Mat{S}$, with all the other parameters fixed, reduces to a linear least-squares problem, which has a closed-form solution.

To start optimization, we initialize $\Mat{V}$ as the PCA subspace of the source data, and take $\Mat{U}$ as its truncated null space. We then obtain the corresponding $\Mat{T}$ and $\Mat{S}$ as described above and start iterating. The pseudo-code of FRODA is provided in Algorithm~\ref{alg:dfroda}. The steps are repeated until convergence, which typically occurs around 50 iterations, with each iteration taking roughly 0.05s.

\begin{algorithm}[!t]
\caption{: FRODA: Factorized Representations for Open-set Domain Adaptation}
\label{alg:dfroda}
\begin{algorithmic}[1]
\vspace{0.1cm}
\REQUIRE
~\\
$\Mat{X}_s \in \mathbb{R}^{D\times n_s}$: the source samples\\
$\Mat{X}_t \in \mathbb{R}^{D\times n_t}$: the target samples\\
$d \ll D$: the dimensionality of the subspaces\\
\vspace{0.1cm}
\ENSURE
~\\
$\Mat{S} \in \mathbb{R}^{d \times n_s}$, $\Mat{T} \in \mathbb{R}^{2d \times n_t}$

~\\
\hspace{-3ex}\textbf{Initialize:} 
~\\
 $\Mat{V} \gets PCA(\Mat{X}_s)$ \\
 $\Mat{U} \gets Null(\Mat{V})$ (i.e., truncated null space of $\Mat{V}$) \\
 
 \noindent
 \STATE Compute $\Mat{T}$ from~\eqref{eqn:eq4} by proximal gradient descent\\
 \STATE Compute $\Mat{S}$ by solving the linear least-squares problem $\min_{\Mat{S}} \|\Mat{X}_s-\Mat{V}\Mat{S}\|_F^2$ \\

\REPEAT
	\STATE Compute $\Mat{U}$ from~\eqref{eqn:eq2} by the Lagrange dual method of~\citep{lee2007efficient}
	\STATE Compute $\Mat{V}$ from~\eqref{eqn:eqv} by the Lagrange dual method of~\citep{lee2007efficient}
	\STATE Compute $\Mat{T}$ from~\eqref{eqn:eq4} by proximal gradient descent
	\STATE Compute $\Mat{S}$ by solving the linear least-squares problem $\min_{\Mat{S}} \|\Mat{X}_s-\Mat{V}\Mat{S}\|_F^2$
\UNTIL{convergence}
\end{algorithmic}
\end{algorithm}

\vspace{-0.2cm}
\paragraph{Inference.} After obtaining the final $\Mat{T}$, we determine if each sample $i$ belongs to known classes or unknown ones based on the coefficients $\Mat{T}_i^u$ and $\Mat{T}_i^v$. More specifically, given a threshold $\varepsilon$, a target sample is assigned to the unknown classes if $ \|\Mat{T}_i^v\| / \|\Mat{T}_i^u\| \leq \varepsilon$, which suggests that it can be well-reconstructed by the private subspace. In the presence of $C$ known classes, we then train a $(C+1)$-way classifier by augmenting the source data with the target samples identified as unknown.


\subsection{D-FRODA: Discriminative FRODA}
The formulation above does not make use of the source labels at all during the representation learning stage. As such, it does not encourage the representation to be discriminative. To overcome this, we extend our basic formulation to further account for the classification task at hand. Specifically, let $\Mat{L}=[\Mat{l}_1 \dots \Mat{l}_{n_s}] \in \mathbb{R}^{C \times n_s}$ be the matrix containing the source labels, where $\Mat{l}_i \in \mathbb{R}^C$ represents the one-hot encoding of the label of sample $i$. We then write our D-FRODA formulation as
\begin{align}
\min_{\Mat{U},\Mat{T},\Mat{V},\Mat{S},\Mat{W}} & \hspace{2ex} \|\Mat{X}_t-\Mat{B}\Mat{T}\|_F^2 +\alpha \|\Mat{X}_s-\Mat{V}\Mat{S}\|_F^2+ \beta \|\Mat{L}-\Mat{W}\Mat{S}\|_F^2 +  \lambda_1 \sum_{i=1}^{n_t}( \|T_i^v\| + \|T_i^u\|))\nonumber\\ 
 s.t.&  \hspace{2ex} \sum_{j=1}^{d}\|\Mat{U}_j\| \leq 1,\sum_{j=1}^{d}\|\Mat{V}_j\| \leq 1\;, &
\label{eqn:eq5}
\end{align}
where $\Mat{W} \in \mathbb{R}^{C \times d}$ is the matrix containing the parameters of a linear classifier for the source data. 

\vspace{-0.2cm}
\paragraph{Optimization.}
To optimize~\eqref{eqn:eq5}, we follow a similar alternating strategy as before. The $\Mat{B}$-minimization and $\Mat{T}$-minimization steps are unchanged, but the $\Mat{S}$-minimization now incorporates a new term and we further need to solve for the classifier parameters $\Mat{W}$. This translates to:

$\Mat{S}$-minimization: Minimizing~\eqref{eqn:eq5} with respect to $\Mat{S}$, with all the other parameters fixed, still reduces to a linear least-squares problem. The two terms involving $\Mat{S}$ can be grouped into a single one of the form $\|\Mat{X}_{new}-\Mat{V}_{new}\Mat{S}\|_F^2$, where $\Mat{X}_{new} = \begin{pmatrix}
\sqrt{\alpha}\Mat{X}_s \\ \sqrt{\beta}\Mat{L} 
\end{pmatrix}$ and $\Mat{V}_{new} = \begin{pmatrix}
\sqrt{\alpha}\Mat{V} \\ \sqrt{\beta}\Mat{W} 
\end{pmatrix}$, and thus $\Mat{S}$ can be obtained in closed form.
\mycomment{
\begin{align}
\min_{\Mat{S}} \hspace{2ex} \|\Mat{X}_{new}-\Mat{V}_{new}\Mat{S}\|_F^2\;,
\end{align}
where $\Mat{X}_{new} = \begin{pmatrix}
\sqrt{\alpha}\Mat{X}_s \\ \sqrt{\beta}\Mat{L} 
\end{pmatrix}$ and $\Mat{V}_{new} = \begin{pmatrix}
\sqrt{\alpha}\Mat{V} \\ \sqrt{\beta}\Mat{W} 
\end{pmatrix}$, and thus $\Mat{S}$ can be obtained in closed form.
}

$\Mat{W}$-minimization: With all the other parameters fixed, finding $\Mat{W}$ corresponds to a linear least-squares problem, with a closed-form solution.

\vspace{-0.2cm}
\paragraph{Inference.}
The same inference strategy as before can be followed to label the target samples. Another option here is to make use of $\Mat{W}$ to classify the samples identified as belonging to known classes. We compare these two strategies in our experiments.

\subsection{D-FRODA-U: D-FRODA with Unknown Source Classes}
Until now, we have tackled the scenario where there are no unknown classes in the source data, which we believe corresponds to the typical application scenario, since the source data can in general be fully annotated. Nevertheless, to match the scenario of~\cite{busto2017open}, that assumes to have access to additional source samples from unknown classes, yet different from the target unknown classes, we introduce a modified version of our approach that takes such auxiliary data into account. Note that, since one knows which source samples are from unknown classes, it is also always possible to simple discard them from training. To nonetheless handle them, we re-write~\eqref{eqn:eq5} as
\begin{align}
\min_{\Mat{U},\Mat{T},\Mat{V},\Mat{U'},\Mat{S'},\Mat{W'}} & \hspace{2ex} \|\Mat{X}_t-\Mat{B}\Mat{T}\|_F^2 +\alpha \|\Mat{X'}_s-\Mat{B'}\Mat{S'}\|_F^2+ \beta \|\Mat{L}-\Mat{W'}\Mat{S'}\|_F^2 \nonumber\\ 
\hspace{-2ex}&  + \lambda_1 \sum_{i=1}^{n_t}\left( \|\Mat{T}_i^v\| + \|\Mat{T}_i^u\|\right) + \lambda_2 \sum_{i=1}^{n_s}\left(\|\Mat{S'}_i^v\| + \|\Mat{S'}_i^u\|\right)\nonumber\\
 s.t.&  \hspace{2ex} \sum_{j=1}^{2d}\|\Mat{B}_j\| \leq 1,\sum_{j=1}^{2d}\|\Mat{B'}_j\| \leq 1\;, &
\label{eqn:eq6}
\end{align}
where $\Mat{X'}_s$ contains the source samples from both known and unknown classes, and $\Mat{B'} = [\Mat{V}, \Mat{U'}] \in \mathbb{R}^{D \times 2d}$ denotes the source transformation matrix with $\Mat{U'} \in \mathbb{R}^{D \times d}$ the private subspace for the source data. Note that, similarly to the target coefficients, we have now separated the source coefficients for each sample $\Mat{S'}_i$ into a part corresponding to the shared subspace $\Mat{S'}_i^v$ and a part corresponding to the private one $\Mat{S'}_i^u$. Note also that the classifier parameters $\Mat{W'}$ now account for $C+1$ classes, the additional class corresponding to the unknown samples.

\vspace{-0.2cm}
\paragraph{Optimization.} We follow a similar iterative procedure to the one used before, with modifications to update $\Mat{B'}$ and $\Mat{S'}$, as discussed below.

$\Mat{S'}$-minimization: To minimize~\eqref{eqn:eq6} with respect to $\Mat{S'}$, with all the other parameters fixed, we write 
\begin{align}
\min_{\Mat{S'}} & \hspace{2ex} \alpha \|\Mat{X'}_s-\Mat{B'}\Mat{S'}\|_F^2+ \beta \|\Mat{L}-\Mat{W'}\Mat{S'}\|_F^2 +  \lambda_2 \sum_{i=1}^{n_s}\left( \|\Mat{S'}_i^v\| + \|\Mat{S'}_i^u\|\right)\;.
\label{eqn:eq7}
\end{align}
The first two terms can be grouped into a single squared Frobenius norm, thus resulting in a sparse group lasso problem, which, as when updating $\Mat{T}$ in FRODA, can be solved via proximal gradient descent~\citep{mairal2014spams}.

$\Mat{B'}$-minimization: Given the coefficients $\Mat{S'}^v$ and $\Mat{S'}^u$, we can update $\Mat{U'}$ by solving
\begin{align}
\min_{\Mat{U'}} & \hspace{2ex} \|\Mat{A}-\Mat{U'}\Mat{S'}^u\|_F^2  \hspace{2ex}
 s.t.   \sum_{j=1}^{d}\|\Mat{U'}_j\|_2 \leq 1\;, 
\label{eqn:eq8}
\end{align}
with $\Mat{A}=\Mat{X'}_s-\Mat{V}\Mat{S'}^v$, and $\Mat{V}$ by solving
\begin{align}
\min_{\Mat{V}} & \hspace{2ex} \|\Mat{A'}-\Mat{V}\Mat{C}\|_F^2  \hspace{2ex}
 s.t.   \sum_{j=1}^{d}\|\Mat{V}_j\|_2 \leq 1 \;, 
\label{eqn:eq9}
\end{align}
with $\Mat{A'} = \begin{pmatrix}
\Mat{X}_t-\Mat{U}\Mat{T}^u\\\sqrt{\alpha}( \Mat{X'}_s-\Mat{U'}\Mat{S'}^u)
\end{pmatrix}$ and $\Mat{C} = \begin{pmatrix}
\Mat{T}^v \\ \sqrt{\alpha}\Mat{S'}^v 
\end{pmatrix}$.
As in FRODA, these two sub-problems can be solved efficiently using the Lagrange dual formulation of~\citep{lee2007efficient}.

\begin{table*}[t!]
				\centering
				     \vspace{1ex}
				
					\begin{tabular}{|l|cccccc|}
						
						\hline
						Method  & $\mathbf{B \rightarrow C}$ & $\mathbf{B \rightarrow I}$ & $\mathbf{B \rightarrow S}$ & $\mathbf{C \rightarrow B}$ & $\mathbf{C\rightarrow I}$ & $\mathbf{C \rightarrow S}$ \\
						\hline
						\hline

TCA~\cite{pan2011domain} & $62.8\pm3.8$ & $56.6\pm4.5$  & $29.6\pm4.2$ & $38.9\pm1.9$ & $60.2\pm1.4$ & $29.7\pm1.6$\\
GFK~\cite{gong2012geodesic} & $66.2\pm4.0$ & $58.3\pm3.1$  & $23.8\pm2.0$ & $40.2\pm1.8$ & $62.2\pm1.5$ & $28.5\pm1.0$ \\
SA~\cite{fernando2013unsupervised}& $66.0\pm3.4$ & $57.8\pm3.2$  & $24.3\pm2.6$ & $40.3\pm1.7$ & $62.5\pm0.8$ & $29.0\pm1.5$\\
CORAL~\cite{sun2016return} & $68.8\pm3.3$ & $60.9\pm2.6$  & $27.2\pm3.9$ & $40.7\pm1.5$ & $64.0\pm2.6$ & $31.4\pm0.8$ \\
ATI~\cite{busto2017open} & $71.4\pm2.3$ & $69.0\pm2.8$  & $37.4\pm2.6$ & $45.7\pm3.0$ & $67.9\pm4.2$ & $37.5\pm2.7$ \\
	\hline
FRODA & $73.8\pm6.1$ & $71.0\pm2.0$  & $54.7 \pm 2.9$ & $67.5\pm1.4$ & $74.5\pm1.7$ & $61.6\pm2.2$ \\
D-FRODA & $\mathbf{74.6\pm5.5}$ & $\mathbf{71.4\pm2.0}$  & $\mathbf{55.4\pm2.7}$ & $\mathbf{67.6\pm1.2}$ & $\mathbf{75.0\pm1.8}$ & $\mathbf{61.7\pm2.1}$\\ \hline
\end{tabular}
\vspace{-3ex}
\end{table*}
  \begin{table*}[t!]
				\centering
				     \vspace{1ex}
				
					\begin{tabular}{|l|cccccc|}
						
						\hline
   					Method & $\mathbf{I \rightarrow B}$ & $\mathbf{I \rightarrow C}$ & $\mathbf{I \rightarrow S}$ & $\mathbf{S \rightarrow B}$ & $\mathbf{S \rightarrow C}$ & $\mathbf{S \rightarrow I}$ \\
					\hline
					\hline						
TCA~\cite{pan2011domain} & $40.9\pm2.9$ & $68.6\pm1.8$  & $34.5\pm3.8$ & $19.4\pm2.1$ & $32.0\pm3.9$ & $31.1\pm4.6$\\
GFK~\cite{gong2012geodesic} & $42.6\pm2.4$ & $73.3\pm3.6$  & $32.7\pm3.6$ & $16.9\pm1.5$ & $28.6\pm3.8$ & $26.4\pm1.1$ \\
SA~\cite{fernando2013unsupervised}& $43.1\pm1.6$ & $72.8\pm3.1$  & $32.2\pm3.7$ & $17.5\pm1.6$ & $29.2\pm4.2$ & $27.1\pm1.3$\\
CORAL~\cite{sun2016return} & $44.6\pm2.5$ & $74.5\pm3.4$  & $35.4\pm4.4$ & $18.7\pm1.2$ & $33.6\pm5.3$ & $31.3\pm1.3$ \\
ATI~\cite{busto2017open} & $48.8\pm2.3$ & $77.5\pm2.2$  & $43.4\pm4.8$ & $23.2\pm3.2$ & $47.3\pm2.9$ & $33.0\pm1.1$ \\
	\hline
FRODA & $66.0\pm1.9$ & $79.9\pm1.7$  & $59.2 \pm 2.1$ & $\mathbf{55.7\pm2.5}$ & $\mathbf{61.2\pm1.8}$ & $59.4\pm1.9$ \\
D-FRODA & $\mathbf{66.4\pm1.7}$ & $\mathbf{80.5\pm1.6}$  & $\mathbf{59.8\pm2.0}$ & $55.5\pm2.4$ & $\mathbf{61.2\pm1.9}$ & $\mathbf{59.6\pm2.2}$ \\

  \hline
					
					\end{tabular}
                    \centering
					\caption {	
						Recognition accuracies on the 12 source/target pairs of the BCIS dataset~\cite{tommasi2014testbed} using a linear SVM classifier. {\bf B}: Bing, {\bf C}: Caltech256, {\bf I}: ImageNet, {\bf S}: SUN.
					}
					\label{tab:dense_cross}
				\end{table*}

\section{Experiments}
\label{sec:experiments}
We evaluate our approach on the task of open-set visual domain adaptation using two benchmark datasets, and compare its performance against the state-of-the-art ATI method of~\cite{busto2017open}\footnote{Among the different variants of this method, we report the top-performing one in each experiment.}, which constitutes the only method designed for the open-set scenario. Note that we also report the results of the methods used as baselines in~\cite{busto2017open}. For a dataset with $C$ source classes, we report the accuracy on $C+1$ classes, the additional one corresponding to the unknown case.

\vspace{-0.2cm}
\paragraph{Implementation details.}
Following~\cite{busto2017open}, we represent the source and target samples with 4096-dimensional $DeCAF_7$ features~\citep{donahue2013decaf} and first reduce their dimensionality by performing PCA jointly on the source and target data and keeping the components encoding 99\% of the data variance. To then determine the dimensionality $d$ of our shared and private subspaces, we make use of the subspace disagreement measure of~\citep{gong2012geodesic}.  For all our experiments, the hyperparameters of our approach were set as follows: $\alpha=0.1$, $\beta=0.01$, $\lambda=0.001$ and $\varepsilon=0.2$. For recognition, for the comparison with~\cite{busto2017open} to be fair, we employ a linear SVM classifier in a one-vs-one fashion. Nevertheless, we also report results obtained with a $k-$Nearest-Neighbor classifier (with $k=3$) and with our linear classifier with parameters $\Mat{W}$ learnt during training.

\paragraph{Results on the dense cross-dataset benchmark.} We first evaluate our approach on the challenging cross-dataset benchmark of~\cite{tommasi2014testbed}. This dataset contains images depicting 40 object categories and coming from four datasets, namely Bing (B), Caltech256 (C), ImageNet (I) and SUN (S), hence referred to as BCIS. Following~\cite{busto2017open}, we consider the samples from the first 10 classes as known instances, while the samples with class labels $11, 12, \cdots, 25$ and $26, 27, \cdots, 40$ are taken to be the unknown samples in the source and target domains, respectively. We follow the unsupervised protocol of~\cite{tommasi2014testbed}, which relies on 50 source samples per class and 30 target images per class, except when the target data is coming from SUN, in which case only 20 images per class are employed.

In Table~\ref{tab:dense_cross}, we compare the results of our methods with those of the baselines on all 12 domain pairs of this dataset. Note that our algorithms (both with and without the discriminative term) significantly outperform all the baselines by a large margin, including the state-of-the-art method of~\cite{busto2017open}. For instance, the margin exceeds 32, resp. 26, percentage points when going from SUN to Bing and ImageNet, respectively. This, we believe, clearly evidences the benefits of our factorized representations, which allow us to separate the unknown target samples from the ones coming from known classes, thus yielding a better representation for the known classes. 

\begin{table*}[t!]
				\centering
				     \vspace{1ex}
				    \scalebox{0.94}{
					\begin{tabular}{|l|cccccc|}
						
						\hline
						Method  & $\mathbf{B \rightarrow C}$ & $\mathbf{B \rightarrow I}$ & $\mathbf{B \rightarrow S}$ & $\mathbf{C \rightarrow B}$ & $\mathbf{C\rightarrow I}$ & $\mathbf{C \rightarrow S}$ \\
						\hline
						\hline
FRODA-SVM & $73.8\pm6.1$ & $71.0\pm2.0$  & $54.7 \pm 2.9$ & $67.5\pm1.4$ & $74.5\pm1.7$ & $61.6\pm2.2$ \\
FRODA-NN & $67.7\pm2.8$ & $64.1\pm2.4$  & $54.4\pm3.5$ & $65.1\pm2.9$ & $72.9\pm1.7$ & $60.5\pm2.0$ \\
	\hline
D-FRODA-SVM & $\mathbf{74.6\pm5.5}$ & $\mathbf{71.4\pm2.0}$  & $55.4\pm2.7$ & $\mathbf{67.6\pm1.2}$ & $\mathbf{75.0\pm1.8}$ & $\mathbf{61.7\pm2.1}$ \\
D-FRODA-W & $61.2\pm1.2$ & $59.3\pm1.1$  & $53.3\pm3.0$ & $63.1\pm0.9$ & $66.2\pm1.7$ & $60.2\pm2.1$ \\
D-FRODA-NN & $67.7\pm3.3$ & $63.3\pm2.8$  & $54.0\pm3.5$ & $65.4\pm2.8$ & $73.4\pm1.5$ & $60.3\pm2.4$ \\
	\hline    
D-FRODA-U-SVM & $71.9\pm3.8$ & $69.2\pm2.7$  & $\mathbf{56.7\pm3.3}$ & $64.8\pm1.8$ & $72.8\pm2.7$ & $59.4\pm2.3$ \\
D-FRODA-U-W & $55.9\pm3.6$ & $55.5\pm4.5$  & $41.7\pm4.8$ & $52.6\pm4.7$ & $61.4\pm3.6$ & $49.5\pm4.1$ \\
D-FRODA-U-NN & $57.6\pm9.1$ & $52.2\pm5.0$  & $47.5\pm6.5$ & $51.8\pm5.7$ & $64.0\pm5.7$ & $57.7\pm5.6$\\ \hline
\end{tabular}}
\vspace{-3ex}
\end{table*}
 \begin{table*}[t!]
				\centering
				  \scalebox{0.94}{
					\begin{tabular}{|l|cccccc|}
						
						\hline
					Method & $\mathbf{I \rightarrow B}$ & $\mathbf{I \rightarrow C}$ & $\mathbf{I \rightarrow S}$ & $\mathbf{S \rightarrow B}$ & $\mathbf{S \rightarrow C}$ & $\mathbf{S \rightarrow I}$ \\
					\hline
					\hline						
FRODA-SVM & $66.0\pm1.9$ & $79.9\pm1.7$  & $59.2 \pm 2.1$ & $55.7\pm2.5$ & $61.2\pm1.8$ & $59.4\pm1.9$ \\
FRODA-NN & $60.9\pm3.7$ & $77.7\pm2.8$  & $58.0\pm2.2$ & $53.4\pm2.2$ & $61.2\pm1.4$ & $58.1\pm1.5$ \\
	\hline
D-FRODA-SVM & $\mathbf{66.4\pm1.7}$ & $\mathbf{80.5\pm1.6}$  & $\mathbf{59.8\pm2.0}$ & $55.5\pm2.4$ & $61.2\pm1.9$ & $59.6\pm2.2$ \\
D-FRODA-W & $62.2\pm1.6$ & $70.5\pm2.5$  & $58.1\pm1.7$ & $\mathbf{56.4\pm1.9}$ & $58.9\pm1.5$ & $58.5\pm0.7$ \\
D-FRODA-NN & $60.9\pm4.1$ & $78.7\pm2.8$  & $57.7\pm2.0$ & $53.0\pm2.3$ & $61.2\pm1.2$ & $57.9\pm1.6$ \\
	\hline    
D-FRODA-U-SVM & $66.0\pm1.2$ & $76.8\pm1.9$  & $57.2\pm4.5$ & $56.3\pm1.9$ & $\mathbf{61.8\pm3.0}$ & $\mathbf{59.9\pm1.7}$ \\
D-FRODA-U-W & $54.6\pm4.4$ & $66.2\pm3.8$  & $43.9\pm5.4$ & $47.6\pm3.8$ & $53.5\pm4.9$ & $51.3\pm3.7$ \\
D-FRODA-U-NN & $58.4\pm6.1$ & $70.9\pm4.5$  & $52.8\pm9.0$ & $55.4\pm2.0$ & $61.5\pm2.0$ & $60.1\pm1.7$ \\
  \hline
					
					\end{tabular}}
					\caption {	
						Recognition accuracies of variants of our approach using linear SVM, nearest neighbor (NN) and our linear classifier ($\Mat{W}$) on the 12 source/target pairs of the BCIS dataset~\cite{tommasi2014testbed}. {\bf B}: Bing, {\bf C}: Caltech256, {\bf I}: ImageNet, {\bf S}: SUN.
					}
					\label{tab:dense_cross_us}
				\end{table*}
				
In Table~\ref{tab:dense_cross_us}, we compare different versions of our method, corresponding to using different classifiers and to using additional unknown source data. Note that the linear SVM classifier, when used with our framework, tends to perform the best, followed by the NN one and finally the learnt linear classifier. This, we believe, can be explained by the fact that, while the linear classifier helps learning a more discriminative representation, it remains less powerful than the other two classifiers to correctly label the target samples. Note also that the use of unknown source data does not consistently help in our framework. Nevertheless, the corresponding results still outperform those of~\cite{busto2017open}.

\begin{wrapfigure}{r}{0.5\textwidth}
 	\vspace{-0.15in}
 	\centering
 	\includegraphics[valign=T, width=7cm]{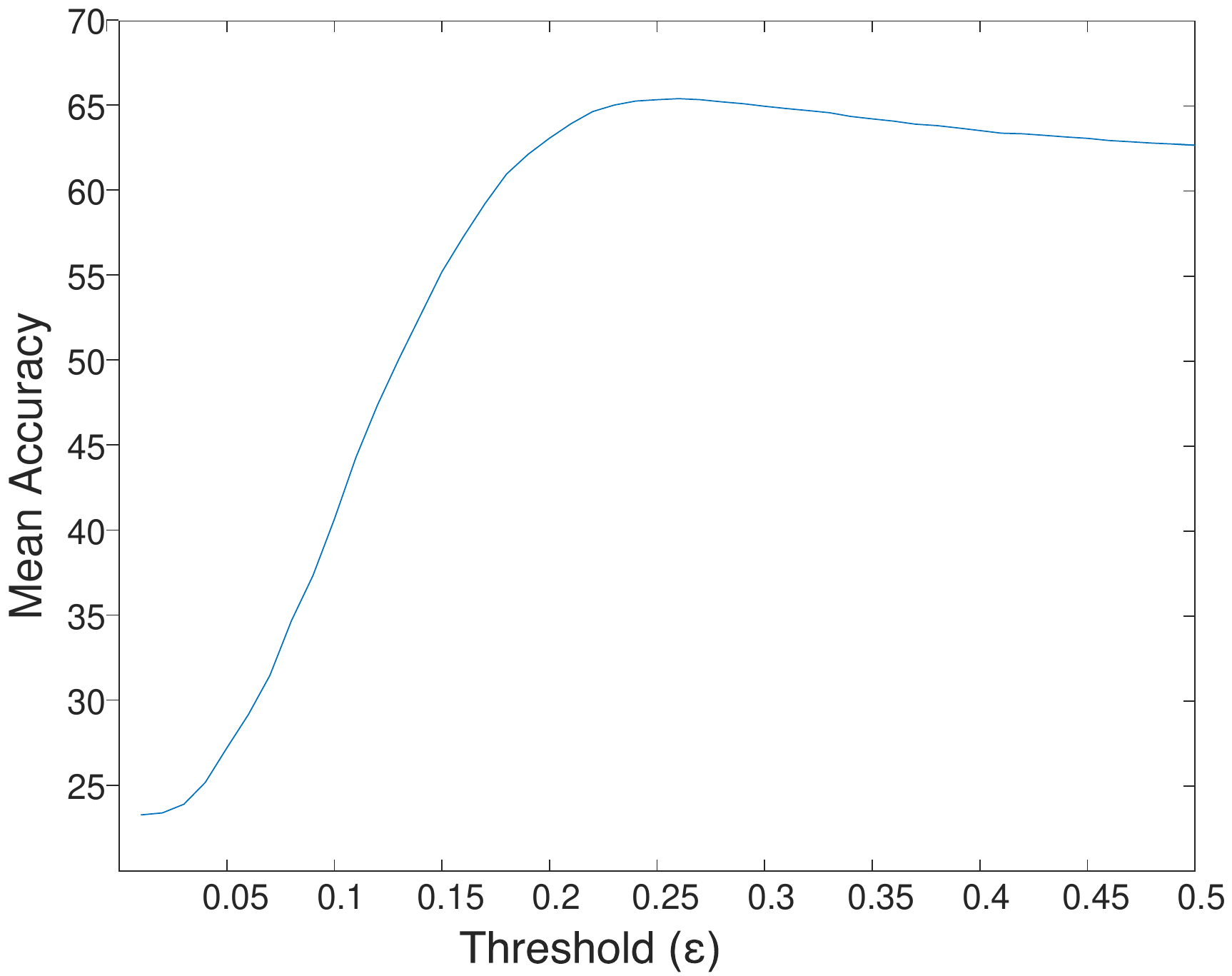}
 	\vspace{-0.05in}
 	\caption{\label{fig:sens} Sensitivity to $\varepsilon$}
 	\vspace{-0.1in}
 \end{wrapfigure}
 
We further evaluate the robustness of our approach to the choice of threshold $\varepsilon$ to separate the target samples from known/unknown classes. In Fig.~\ref{fig:sens}, we plot the average accuracy over all 12 pairs of the BCIS dataset as a function of the value of $\varepsilon$. Note that, once a sufficiently large threshold is reached, the results are quite stable. This indicates that our algorithm is robust to the specific value of this hyperparameter.


\vspace{-0.2cm}
\paragraph{Results on the Office dataset.} We further evaluate our approach on the slightly less challenging, although standard Office benchmark~\citep{saenko2010adapting}. This dataset contains three different domains, namely Amazon (A), DSLR (D) and Webcam (W), sharing 31 object categories, but differing in data acquisition process. As in~\cite{busto2017open}, we take all the samples from the first 10 classes to represent the known ones, and all the samples with class labels $11, 12, \cdots , 20$ and $21, 22, \cdots, 31$ as unknown source and target data, respectively.

We report the results of our algorithms and of the baselines for all 6 domain pairs of this dataset in Table~\ref{tab:office_results}. As before, note that we outperform the baselines in this open-set scenario. In Table~\ref{tab:office_results_us}, we compare the different variants of our approach. The conclusions that one can draw from these results are similar to those for the BCIS dataset, thus showing that our method generalizes well across different domain adaptation datasets.

\vspace{-0.2cm}
\paragraph{Runtimes.} As mentioned in Section~\ref{sec:proposed}, one iteration of our approach takes on average 0.05 second, and our algorithm typically takes around 50 iterations to converge. This yields a total runtime of roughly 2.5 seconds. By contrast, the publicly available implementation of the method of~\cite{busto2017open} takes on average 8 seconds per iteration and typically converges in 4 iterations, thus corresponding to a total runtime of roughly 32 seconds. Note that these runtimes were computed on the same computer. Therefore, not only is our approach significantly more accurate than that of~\cite{busto2017open}, but it also is faster by one order of magnitude.

				\begin{table*}[t!]
				\centering
				     \vspace{1ex}
				    \scalebox{1.1}{
					\begin{tabular}{|l|cccccc|}
						
						\hline
						Method  & $\mathbf{A\rightarrow D}$ & $\mathbf{A\rightarrow W}$ & $\mathbf{W\rightarrow A}$ & $\mathbf{W\rightarrow D}$ & $\mathbf{D\rightarrow A}$ & $\mathbf{D\rightarrow W}$  \\
						\hline
						\hline			
						
LSVM & $72.6$ & $57.5$ & $49.2$ &  $98.8$ & $45.1$ & $88.5$ \\				
DAN~\cite{long2016deep} & $77.6$ & $72.5$ & $60.8$ & $98.3$ & $57$ & $88.4$ \\				
RTN~\cite{long2016unsupervised} & $76.6$ & $73$ & $62.4$ & $98.8$ & $57.2$ &  $89$  \\						
BP~\cite{backprop} & $78.3$ & $75.9$ & $64$ & $98.7$ & $57.6$ &  $89.8$ \\
ATI~\cite{busto2017open} & $79.8$ & $78.4$ & $76.7$ & $\mathbf{98.8}$ & $71.3$ & $94.4$  \\
						
						\hline
						\hline
FRODA & $\mathbf{88.0}$ & $\mathbf{78.7}$ & $76.5$ & $98.0$ & $\mathbf{73.7}$ & $\mathbf{94.6}$ \\
D-FRODA & $87.4$ & $78.1$ & $\mathbf{77.1}$ & $98.5$ & $73.6$ & $94.4$  \\
\hline
					\end{tabular}}
					\caption {	
						Recognition accuracies on the 6 source/target pairs of the Office dataset~\cite{saenko2010adapting} using a linear SVM classifier. {\bf A}: Amazon, {\bf W}: Webcam, {\bf D}: DSLR.
					}
					\label{tab:office_results}
				\end{table*}

\begin{table*}[t!]
				\centering
				     \vspace{1ex}
				     \scalebox{1.05}{
					\begin{tabular}{|l|cccccc|}
						
						\hline
Method  & $\mathbf{A\rightarrow D}$ & $\mathbf{A\rightarrow W}$ & $\mathbf{W\rightarrow A}$ & $\mathbf{W\rightarrow D}$ & $\mathbf{D\rightarrow A}$ & $\mathbf{D\rightarrow W}$  \\
						\hline
						\hline			
						
FRODA-SVM & $\mathbf{88.0}$ & $78.7$ & $76.5$ & $98.0$ & $\mathbf{73.7}$ & $\mathbf{94.6}$ \\
FRODA-NN & $83.9$ & $69.5$ & $75.0$ &  $97.7$ & $69.0$ & $83.9$ \\
 						\hline
 						\hline
D-FRODA-SVM & $87.4$ & $78.1$ & $\mathbf{77.1}$ & $\mathbf{98.5}$ & $73.6$ & $94.4$  \\
D-FRODA-NN & $83.9$ & $70.1$ & $75.1$ & $96.8$ & $69.2$ & $84.5$ \\
D-FRODA-W & $71.1$ & $65.3$ & $68.1$ & $83.0$ & $67.3$ & $79.1$ \\
 						\hline
 						\hline
D-FRODA-U-SVM& $81.9$ & $\mathbf{83.5}$ & $75.5$ & $96.2$ & $70.6$ & $94.2$ \\
D-FRODA-U-NN & $78.1$ & $72.1$ & $69.1$ & $93.6$ & $67.0$ & $75.7$ \\
D-FRODA-U-W & $73.4$ & $65.9$ & $62.8$ & $88.0$ & $61.7$ & $65.1$  \\
\hline
 					\end{tabular}}
					\caption {	
						Recognition accuracies of variants of our approaches using a linear SVM, nearest neighbor (NN) and our linear classifier ($\Mat{W}$) on the 6 source/target pairs of the Office dataset~\cite{saenko2010adapting}. {\bf A}: Amazon, {\bf W}: Webcam, {\bf D}: DSLR.
					}
					\label{tab:office_results_us}
\end{table*}

\section{Conclusion}
\label{sec:conclusion}
We have introduced a novel approach to open-set domain adaptation, based on the intuition that source and target samples coming from the same, known classes can be represented by a shared subspace, while target samples from unknown classes should be modeled with a private subspace. Each step of the resulting algorithms can be solved efficiently. Therefore, as demonstrated by our experiments, our method outperforms the state of the art in open-set domain adaptation not only in terms of accuracy, but also in terms of speed, being one order of magnitude faster. We believe that this clearly evidences the benefits of learning factorized representations, which allows us to jointly discard the unknown target samples and learn a better shared representation. In the future, we will investigate ways to make better use of unknown source data, and to exploit more effective classifiers, such as SVM, directly within our D-FRODA formulation.

\bibliographystyle{ieee}
\bibliography{references}

\end{document}